\documentclass[11pt]{article}

\usepackage[final]{acl}

\usepackage{times}
\usepackage{latexsym}
\usepackage[dvipsnames]{xcolor}
\usepackage{amsmath}
\usepackage{amssymb}

\usepackage[T1]{fontenc}
\usepackage[utf8]{inputenc}

\usepackage{microtype}

\usepackage{inconsolata}

\usepackage{graphicx}
\usepackage{caption}
\usepackage{booktabs}

\title{Towards Mitigating Hallucinations in Large Vision-Language Models by Refining Textual Embeddings}

\author{
 \textbf{Aakriti Agrawal \textsuperscript{1}},
  \textbf{Gouthaman KV
  \textsuperscript{2}},
 \textbf{Rohith Aralikatti \textsuperscript{1}},
 \textbf{Gauri Jagatap 
 \textsuperscript{2}}, \\
 \textbf{Jiaxin Yuan 
 \textsuperscript{1}},
  \textbf{Sarvesh Baskar 
 \textsuperscript{1}},
 \textbf{Vijay Kamarshi 
 \textsuperscript{2}},
 \textbf{Andrea Fanelli
 \textsuperscript{2}},
 \textbf{Furong Huang
 \textsuperscript{1,3}}
\\
 \textsuperscript{1} University of Maryland,
 \textsuperscript{2} Dolby Laboratories,
 \textsuperscript{3} Capital One
\\
 \small{
   \textbf{Correspondence:} \href{mailto:agrawal5@umd.edu}{agrawal5@umd.edu}
 }
}

\begin{document}
\maketitle
\begin{abstract}
Hallucinations in Large Vision-Language Models (LVLMs) remain a persistent challenge, often stemming from inadequate integration of visual information during multimodal reasoning. A key cause is the model’s over-reliance on textual priors and underutilization of visual cues, leading to outputs that are linguistically fluent but visually inaccurate. For example, given an image of an empty kitchen countertop, an LVLM might hallucinate a “bowl of fruit” or “cup of coffee,” relying on language associations rather than visual evidence. Most LVLMs incorporate visual features by appending them to the input stream of a pre-trained LLM and training on large-scale vision-language datasets. Our systematic analysis reveals that this strategy often leads to over-dependence on textual information due to the inherent bias of LLMs towards language-dominant representations. This imbalance skews attention towards the text over visual content, weakening the model’s ability to ground outputs in visual inputs. To address this, we propose a simple yet effective visual feature incorporation method that encourages the model to learn visually-informed textual embeddings distinct from those of the base LLM and promotes a more balanced attention distribution. Experimental results across multiple hallucination benchmarks demonstrate that our method significantly reduces hallucinations and fosters more balanced multimodal reasoning. Notably, our approach achieves substantial gains, including \textbf{+9.33\%} on MMVP-MLLM, \textbf{+2.99\%} on POPE-AOKVQA, up to \textbf{+3.4\%} on Merlin, and \textbf{+3\%} 
on the hard-data split of HallusionBench. Code: \href{https://github.com/Aakriti05/VisAlign.git}{https://github.com/Aakriti05/VisAlign.git} 
\end{abstract}
    
% \vspace{-5mm}
\section{Introduction}
\label{sec:intro}
% \vspace{-2mm}

The advent of LLMs has transformed tasks like machine translation, dialogue, and content generation with unprecedented accuracy and fluency. Building on this, Large Vision-Language Models (LVLMs) \citep{lin2023video, zhang2023video, Maaz2023VideoChatGPT} integrate visual and linguistic understanding in a unified framework, bridging text and visual modalities. This synergy has advanced tasks such as captioning~\citep{chen2022pali}, question-answering~\citep{li2023blip}, multimodal retrieval~\citep{lin2024nvmmembed}, etc. As LVLMs advance, their adoption in domains such as healthcare, autonomous driving, and education is revolutionizing real-world AI application.

\begin{figure}[t]%[htp]
    \centering
    \includegraphics[width=\columnwidth, trim=150 150 130 60, clip]{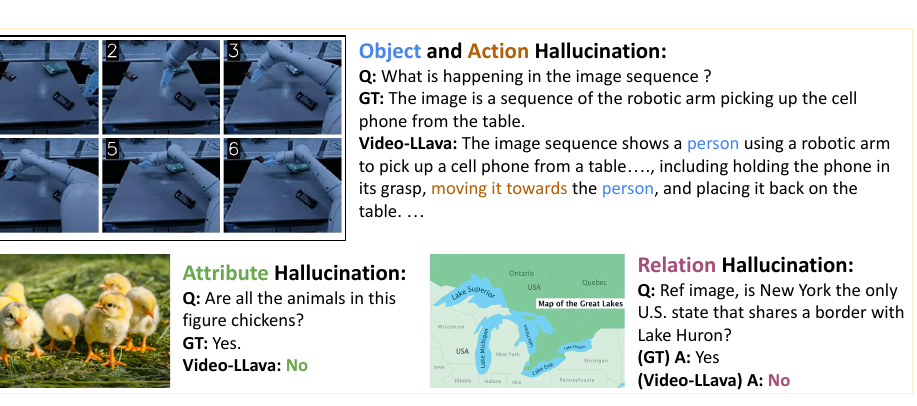}
    % \vspace{-1.8em}
    \small
    \caption{Hallucinations in Video-LLaVA \citep{lin2023video}.} 
    \label{fig:introex}
     % \vspace{-0.5em}
\end{figure}

% \begin{wrapfigure}{r}{0.52\textwidth}
%     \vspace{-1.6em}
%     \centering
%     \includegraphics[width=0.50\textwidth, trim=150 150 130 60, clip]{wacv-2026/figures/hallucination_Examples.pdf}
%     \vspace{-0.8em}
%     \caption{\small Hallucinations in Video-LLaVA \citep{lin2023video}.}
%     \label{fig:introex}
%     \vspace{-1em}
% \end{wrapfigure}

Despite this progress, LVLMs remain prone to hallucinations—outputs that are fluent but not grounded in the visual input. These errors, which include fabricating or misinterpreting visual content, undermine reliability and hinder deployment in safety-critical settings. Fig.~\ref{fig:introex} illustrates failure cases from a LVLM, Video-LLaVA~\citep{lin2023video}. In one example, the model captions a scene as “moving it towards a person,” despite the absence of both the person and the action in the video, demonstrating simultaneous object and action hallucination. More broadly, LVLM hallucinations manifest in several forms: attribute hallucinations, where incorrect visual properties are assigned (e.g., describing a blue car as red or denying visible objects); relation hallucinations, which fabricate spatial or contextual relationships (e.g., claiming a person is jumping over a fence when they are standing beside it); and, in video settings, temporal hallucinations, where nonexistent dynamics are inferred (e.g., asserting that a person enters the room when no such event occurs).

\begin{figure*}%[t]
    \centering
    \small
\includegraphics[width=0.95\textwidth, trim=10 60 10 75, clip]{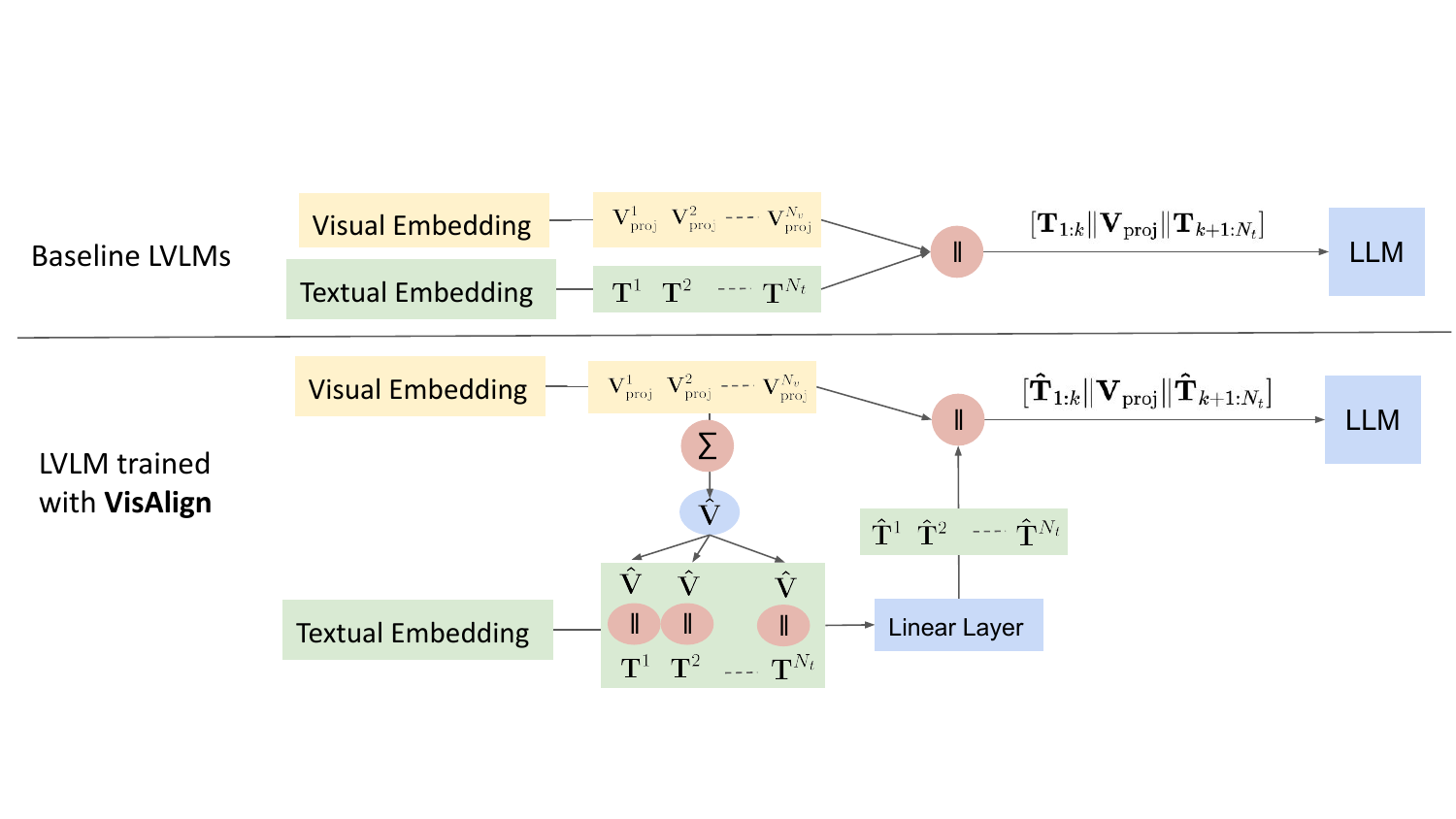}
% \vspace{-1em}
    \caption{
\textbf{Top:} Architecture of typical LVLMs like Video-LLaVA, which fuse language and vision embeddings by simple concatenation. 
\textbf{Bottom:} Our modified architecture with a \textbf{concatenation block that appends the averaged vision embedding to each token embedding, followed by a projection layer}. This encourages the model to learn visually informed textual embeddings and better attend to visual input during training.
}
    \label{fig:introduction}
    % \vspace{-1.5em}

\end{figure*}

We hypothesize that a fundamental source of hallucinations in LVLMs arises from the prevailing architectural paradigm in which the visual information is appended as embeddings to the textual embeddings of a pre-trained LLM (Fig.~\ref{fig:introduction}, top). This fused input is then passed to the model and fine-tuned on large-scale vision-language datasets, such as image/video captioning, and Visual Question Answer (VQA) \citep{lin2023video, he2024efficient, Maaz2023VideoChatGPT}, etc. While this approach offers modularity, data efficiency, and leverages the strong language generation capabilities of LLMs, it introduces a structural asymmetry: the LLM backbone, trained solely on text, remains inherently biased toward language-driven reasoning \citep{an2025mitigating, arif2025paint}. As a result, during fine-tuning, the model may tend to fall back to text priors, under-utilizing the visual embeddings and treating them as secondary in the reasoning process. This modality imbalance may lead to a systematic misalignment between visual evidence and generated text, manifesting during inference as hallucinations: outputs that are linguistically coherent and semantically plausible, yet factually incorrect or unsupported by the visual input.

Motivated by this, we systematically investigate modality imbalance in LVLMs as a potential source of hallucinations, with a focus on the dominant practice of appending visual embeddings to the input textual tokens of pre-trained LLMs~\citep{lin2023video, he2024efficient, Maaz2023VideoChatGPT}. We use Video-LLaVA~\citep{lin2023video} as the main model for study this imbalance due to its strong performance and community adoption. We also show results on LLaVA1.5~\citep{llava1.5} and Open-Qwen2VL~\citep{openqwen2vl} to show generalization ability. 
Our analysis reveals that the prevailing approach of simply appending visual embeddings to the textual input sequence causes the model to over-rely on language while under-utilizing visual information, thereby exacerbating hallucinations. This arises because the backbone LLM, optimized for text, disproportionately emphasizes textual tokens during self-attention operations within the transformer layers. 

To address this modality imbalance, we propose a method that integrates visual information directly into text embeddings at the input token level, enabling more balanced attention and cross-modal representations that better ground generation in visual input. By embedding visual semantics within the language representation, our method fosters balanced cross-modal reasoning, and reduces hallucinations. Extensive evaluation across multiple hallucination benchmarks demonstrates consistent and statistically significant gains, highlighting the effectiveness and generalizability of our method.

% added for ACL
% To address this modality imbalance, we introduce VisAlign that integrates visual information into textual embeddings at the token level, thus redefining the input text representations and enabling joint learning of textual and visual information during training. Extensive evaluations across multiple hallucination benchmarks show consistent and statistically significant improvements, demonstrating the effectiveness and generalizability of our approach. 
%These results further confirm that embedding visual context directly into textual representations encourages the model to rely more strongly on visual evidence.

% the LLM is no longer able to treat text as the primary modality;
% attention becomes more uniformly distributed across modalities, reducing reliance on linguistic shortcuts.
% Collectively, 

\vspace{-2mm}
\section{Related Works}
\vspace{-2mm}
\label{related}
\textbf{LVLMs} extend pre-trained LLMs to handle visual inputs, typically by appending visual embeddings---extracted from frozen image or video encoders---to the language token sequence. This token-level fusion strategy enables architectural modularity and reusability of LLMs without major modifications. Notable models following this approach include LLaVA~\citep{llava_cvpr}, MiniGPT-4~\citep{zhu2023minigpt}, Video-LLaVA~\citep{lin2023video}, Video-ChatGPT~\citep{maaz2023video}, Bunny~\citep{he2024efficient}, Open-Qwen2VL \cite{openqwen2vl} and Video-LLaMA~\citep{zhang2023video}. Among these, Video-LLaVA has strong benchmark performance, open-source, and straightforward temporal extension via frame-wise token concatenation~\citep{tang2025video}. %Tang et al.~\citep{tang2025video} further identify Video-LLaVA as a key reference model that underpins many derivatives, including Video-ChatGPT (chat applications), Bunny (efficiency), and Video-LLaMA (fine-grained fusion). 
Open-Qwen2VL~\citep{openqwen2vl} is also a
fully open-source multimodal model with SOTA performance which instead of concatenating visual tokens, it fuses it directly into the token embedding space, enabling richer cross-modal interactions and stronger native visual grounding. Models like Flamingo~\citep{alayrac2022flamingo} and BLIP-2~\citep{li2023blip} use complex cross-attention to integrate modalities dynamically across transformer layers. Although more flexible, they incur higher computational costs and less modularity. Empirical results~\citep{llava_cvpr} show simpler token-appending strategies often match or outperform these methods in accuracy and efficiency. For its simplicity, extensibility, and strong performance, we adopt Video-LLaVA~\citep{lin2023video} as the primary model to investigate visual feature integration limitations, focusing on attention distribution, modality alignment, and hallucination. We extend main results to LLaVA1.5 and Open-Qwen2VL to show generalization ability.

\textbf{Hallucination Detection and Mitigation in LVLMs} 
Several approaches have recently been proposed to mitigate hallucinations in LVLMs. M-HalDetect~\citep{gunjal2023mhaldetect} introduces a dataset of hallucinated captions for training classifiers, while HaELM~\citep{wang2023haelm} proposes a fine-tuning framework to distinguish hallucinated from faithful outputs. Reinforcement learning methods such as GAVIE~\citep{liu2023gavie} penalize ungrounded generations, and ALOHa~\citep{petryk2023aloha} leverages LLMs to detect hallucinated objects beyond fixed vocabularies. RLHF-based techniques~\citep{sun2023mmhal} further enhance multimodal alignment. CLOCK~\citep{biten2022clock} uses attention calibration during training.
Inference-time strategies include visual-grounding-enhanced decoding via image descriptions~\citep{ghosh2024visual}, Instruction Contrastive Decoding (ICD)~\citep{wang2024icd}, Self-Introspective Decoding (SID)~\citep{huo2024sid}, which verifies partial generations, and Visual Contrastive Decoding (VCD)~\citep{leng2024mitigating}, which re-ranks outputs to promote visual consistency, ClearSight~\citep{yin2025clearsight} and other attention aligning methods \citep{zhao2025mitigating, fazli2025mitigating, jiang2025devils}. Together, these methods represent the current state of the art in hallucination mitigation.

% Unlike prior approaches that rely during inference-time heuristics, or hallucination-supervised fine-tuning, our method is more principled and addresses hallucination proactively at the input representation during training time. 

% added for ACL
Unlike prior approaches that mitigate hallucinations through inference-time decoding strategies or supervised fine-tuning, VisAlign addresses hallucinations at the representation level by refining textual embeddings with global visual context before transformer processing. This shifts multimodal alignment from the attention stage to the embedding stage, enabling more balanced cross-modal reasoning without modifying the LLM architecture.

% Unlike prior approaches that rely on decoding strategies or fine-tuning, VisAlign mitigates hallucinations at the representation level by conditioning textual embeddings on global visual context before transformer processing. This moves multimodal alignment from attention to the embedding stage, enabling more balanced cross-modal reasoning without modifying the LLM architecture.

% By enriching textual embeddings with visual information, we encourage more balanced cross-modal attention and more effective utilization of visual cues---directly targeting the root cause of modality imbalance. This results in a more integrated, principled, and generalizable solution. Furthermore, our method not only performs well independently but also enhances existing techniques, as demonstrated in Appendix~\ref{appendix}, underscoring its broad applicability and complementary strengths.

% added for ACL camera-ready to incorporate reviewers comments
\textbf{Representation-Level Visual Grounding}
Our work is related to a broad class of approaches that incorporate visual information into language representations to improve multimodal grounding. Early fusion and visual-context injection methods integrate visual features into textual representations, while co-attention and cross-modal attention architectures such as LXMERT~\cite{lxmert} and ViLBERT~\cite{vilbert} enable deeper interactions between visual and linguistic features. More recent vision-language pretraining approaches, including CLIP~\cite{clip} and BLIP~\cite{blip}, learn aligned multimodal representations through large-scale image–text supervision. Complementary to these, techniques such as FiLM~\cite{film} perform feature-wise modulation to condition one modality on another at the representation level. In visual question answering, methods such as the Visually-Grounded Question Encoder (VGQE)~\cite{vgqe} and related architectures inject visual features directly into question representations to mitigate language bias and promote early grounding.
In line with these approaches, our work leverages visual information to shape language representations, but focuses specifically on mitigating hallucination in LVLMs by addressing modality imbalance at the representation level. We introduce a lightweight mechanism that aligns textual embeddings with global visual context, improving visual consistency during generation.

\vspace{-3mm}
\section{Background}\label{sec:label}
\vspace{-3mm}

As noted above, we adopt the widely used and open-source Video-LLaVA as our baseline for major experiments but also show performance on LLaVA 1.5 and Open-Qwen2VL to show generalisability of our approach. 
%due to its pivotal role in advancing the field. Its canonical status and strong performance make it an ideal foundation for investigating modality imbalance, attention dynamics, and hallucination behaviors in LVLMs.%~\footnote{Results on additional baseline are provided in the appendix}.
% Given its canonical status and strong performance, Video-LLaVA-7B serves as an ideal baseline for our investigation into modality imbalance, attention dynamics, and hallucination behaviors in LVLMs.% 
Therefore, this section formally outlines the architecture and training pipeline of Video-LLaVA (refer Figure~\ref{fig:introduction} for an overview). It consists of the following components: \\
\noindent \textbf{\textit{A frozen visual encoder}} to extract embeddings from the video (or image), the Video-LLaVA uses the pre-trained LanguageBind~\citep{zhu2023languagebind}. \\
\noindent \textbf{\textit{A projection layer}} that maps the visual embeddings into the textual (base LLM's) embedding space. The vision-language alignment is carried out via this projection layer.  Formally, let $\mathbf{V} \in \mathbb{R}^{N_v \times d_v}$ denote the visual embeddings, where $N_v$ is the number of visual tokens 
%(e.g., image patches or video frames) 
and $d_v$ is the visual embedding dimension. Output from the learnable projection layer
    $\mathbf{W}_p \in \mathbb{R}^{d_v \times d_t}$ is denoted as:

{\small
\begin{equation}    
\mathbf{V}_{\text{proj}} = \mathbf{V} \mathbf{W}_p, \quad \text{where } \mathbf{V}_{\text{proj}} \in \mathbb{R}^{N_v \times d_t}
\label{eq:concat}
\end{equation}}
where $d_t$ is the LLM embedding dimension. \\
\noindent \textbf{\textit{A backbone LLM: }} LVLMs extend upon a pre-trained LLM. Video-LLaVA uses the pre-trained Vicuna-7b \citep{zheng2023judging}. \\
The training consists of two stages: \\
\noindent\textit{\textbf{Pretraining:}} The visual encoder is frozen, and only the projection layer $\mathbf{W}_p$ is trained to align visual embeddings with the LLM’s input space. \\
\noindent\textit{\textbf{Finetuning:}} The full model, including the LLM, is trained end-to-end to enable effective reasoning over combined visual and textual inputs for visually grounded generation tasks.

\vspace{-3mm}
\section{Evaluating Attention Score Distribution}
\vspace{-3mm}

Analyzing the attention score distribution across transformer layers provides insight into how information flows from lower to higher layers in LLMs. These scores reveal which tokens most influence the model's output and offer insight into its learning dynamics~\citep{zhang2023tell}. Extending this analysis to LVLMs, we visualize the attention score distributions over both textual and visual tokens to better understand cross-modal interactions.

\begin{figure*}%[htpb!]
\centering
\includegraphics[width=0.9\textwidth, trim=0 130 0 10, clip]{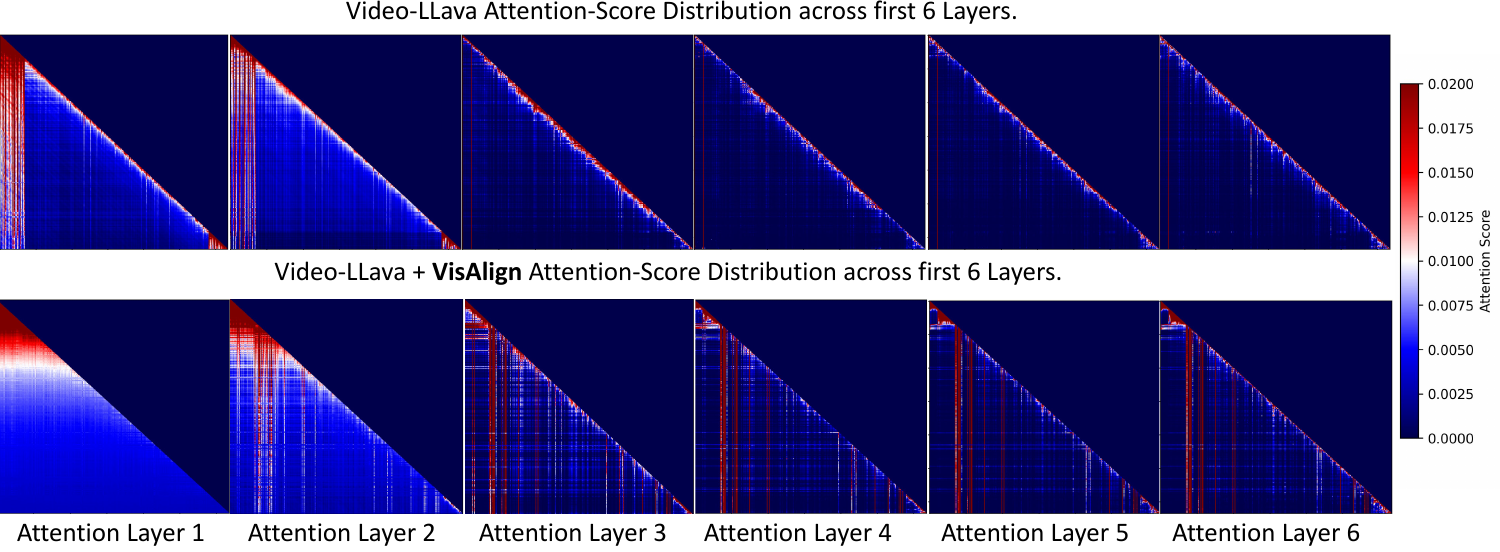}
\vspace{-0.8em}
\caption{Attention score distributions across the first six attention layers of the baseline Video-LLaVA model (top row) and the VisAlign-enhanced model (bottom row). Video-LLaVA concatenates tokens in a fixed order: 35 initial text tokens, followed by 256 visual embeddings, and then the remaining text tokens. In each map, the x-axis denotes attended tokens (keys), and the y-axis denotes attending tokens (queries). Color intensity reflects attention weight: \textcolor{blue}{blue} indicates low attention, \textcolor{red}{red}/white indicates high attention, and dark (near-black) regions indicate masked or negligible attention due to causal masking in autoregressive LVLMs.
}
% \vspace{-2.5em}
\label{fig:attention_layers}
\end{figure*}

Figure~\ref{fig:attention_layers} shows attention score distributions across multiple transformer layers in Video-LLaVA. In each heatmap, the horizontal axis represents Key tokens (tokens being attended to), and the vertical axis represents Query tokens (tokens performing attention). Color intensity encodes attention strength: cooler tones (e.g., \textcolor{blue}{blue}) indicate lower scores, while warmer tones (e.g., \textcolor{red}{red}) and white indicate stronger attention. Nearly black regions show zero attention due to causal masking. This visualization qualitatively reveals how attention is distributed between visual and textual tokens across the network. Asymmetric or modality-skewed patterns highlight if the model overly favors one modality (typically text) at the expense of the other modality (visual), which can explain hallucination and grounding failures in multimodal tasks.

Figure~\ref{fig:attention_layers} reveals a pronounced imbalance in how Video-LLaVA distributes attention between textual and visual tokens. In Layer 1 (top row, first plot), attention is heavily concentrated on the initial textual tokens (upper-left red region), sharply declines over the visual tokens, and rises again for the trailing textual tokens---a pattern consistent across layers. As defined in Eq.~\eqref{eq:concat}, the input sequence $X$ follows a fixed order: initial textual tokens, followed by visual tokens, and then remaining textual tokens. This results in the model disproportionately focusing on textual tokens at both ends while under-attending to the visual tokens in between.

This asymmetric attention distribution reflects a modality bias rooted in the pre-trained base LLM, which was trained exclusively on text. During fine-tuning, the model relies heavily on linguistic priors and insufficiently leverages the visual embeddings provided by the frozen image or video encoder. This imbalance restricts the effective propagation and integration of visual signals across transformer layers, undermining robust visual grounding. Consequently, the model is prone to generate hallucinations---outputs that are fluent and semantically coherent but factually misaligned 
or unsupported by the visual input.

% A quantitative assessment of this visual and textual contribution imbalance is provided in Appendix Section \ref{appx:quant_vis_contri}.
% \vspace{-2mm}
\section{Improving Attention Score Distribution by Refining Textual Embeddings}
% \vspace{-2mm}

We propose a simple yet principled approach, \textbf{VisAlign}, aimed at improving the distribution of attention scores across visual and textual modalities. The underlying hypothesis is that encouraging a more balanced attention pattern—particularly by increasing attention to visual tokens—enables the model to better utilize visual information and reduces hallucinations caused by over-reliance on textual priors. VisAlign operates by refining textual embeddings through the integration of visual context prior to their input to the LLM. This encourages the model to jointly attend to and learn from both textual and visual information during training, leading to more meaningful visual encoding. By fostering a balanced and synergistic interaction between vision and language, VisAlign improves visual utilization without requiring architectural changes or external supervision.

As illustrated in Figure~\ref{fig:introduction}, VisAlign first applies average pooling on the projected visual embeddings $\mathbf{V}_{\text{proj}} \in \mathbb{R}^{N_v \times d_t}$, resulting in the visual embedding vector $\mathbf{\hat{V}} \in \mathbb{R}^{1 \times d_t}$: 

\begin{equation}
\small
\mathbf{\hat{V}} = \frac{1}{N_v} \sum_{m=0}^{m=(N_v-1)} \mathbf{V}_{\text{proj}}[m] 
\end{equation}
% added for ACL
We use average pooling to derive a global visual summary because it provides a stable and parameter-free aggregation of visual tokens with minimal computational overhead. This lightweight design allows us to incorporate global visual context into textual embeddings while focusing on the core goal of VisAlign: evaluating whether embedding-level visual grounding improves multimodal attention and reduces hallucinations.

% We adopt average pooling as a stable and parameter-free aggregation of visual tokens with minimal computational overhead. This lightweight design enables incorporation of global visual context into textual embeddings while isolating the effect of embedding-level visual grounding in reducing hallucinations.

Next, we fuse $\mathbf{\hat{V}}$ with the text embeddings $\mathbf{T}_{1:N_t} \in \mathbb{R}^{N_t \times d_t}$ via concatenation along the $d_t$ dimension, yielding the fused embeddings
$\mathbf{T_V}$:
\vspace{-2mm}
\begin{equation}
\small
\mathbf{T_V} = \left[\, \mathbf{T} \,\middle\|\, \mathbf{\hat{V}} \otimes \mathbf{1}_{N_t} \,\right] \in \mathbb{R}^{N_t \times 2d_t}
\label{eq:text_vision_concat}
\end{equation}
Then, we apply a linear projection layer $\mathbf{W}_d \in \mathbb{R}^{2d_t \times d_t}$ to map the fused representations $\mathbf{T_V}$ back to the original LLM embedding dimension $d_t$, producing the visually-grounded text token sequence,  $\mathbf{\hat{T}} = \mathbf{T_V} \mathbf{W}_d$ ($\in \mathbb{R}^{N_t \times d_t}$). Unlike the original textual tokens which are derived solely from language embeddings,  $\mathbf{\hat{T}}$ is now a modified version of those language embedding still carrying the textual information. This enforces the model to learn these new embeddings like it tries to learn visual embeddings, thus giving better attention distribution and more effective cross-modal reasoning and visual grounding in downstream tasks. Finally, we append $\mathbf{\hat{T}}$ to $\mathbf{V_{proj}}$ following the original concatenation strategy in Video-LLaVA (Eq.~\eqref{eq:concat}):
\begin{equation}
\small
\mathbf{\hat{X}} = [\, \mathbf{\hat{T}}_{1:k} \,\|\, \mathbf{V_{proj}} \,\|\, \mathbf{\hat{T}}_{k+1:N_t} \,] ; \text{where } \mathbf{\hat{X}} \in \mathbb{R}^{(N_t + N_v) \times d_t}
\end{equation}
The token sequence $\mathbf{\hat{X}}$ is then fed into the base LLM. It consists of visually grounded textual embeddings, followed by visual embeddings, and ends with the remaining grounded textual tokens. 
% % added for ACL
Unlike standard token concatenation approaches used in LVLMs, VisAlign injects visual context directly into each textual token embedding, ensuring that visual information is present before self-attention operations in the transformer.

\textbf{Training Stages:}
% \vspace{-1mm}
We use the same datasets and training strategy as used in the baseline VideoLLaVA~\citep{lin2023video} (also discussed in Sec \ref{sec:label}). In the \textit{pretraining stage,} we train both the vision-language projection layer and the linear layer, while keeping the LLM frozen (refer to Fig. \ref{fig:introduction} for an overview). Whereas in the \textit{finetuning stage,} we train the full model end-to-end, including LLM.

% \vspace{-2mm}
\subsection{Attention Score Distribution with VisAlign}
% \vspace{-1.5mm}

 Figure~\ref{fig:attention_layers} (bottom row) shows the attention distribution of Video-LLaVA trained with the VisAlign method. As illustrated, attention with VisAlign is more balanced and structured, spanning both visual and textual tokens throughout the sequence. Notably, the vertical attention bands are sharper and more frequent, indicating that the model consistently attends to specific visual regions or tokens that serve as semantic anchors across layers. Additionally, the smoother and more continuous diagonal gradients indicate that tokens attend not only to their local context but also capture long-range dependencies, reflecting a balanced and context-aware attention mechanism. In contrast, the top row (baseline Video-LLaVA) shows less coherent, more fragmented attention patterns. High attention is concentrated at the sequence boundaries, corresponding to textual token positions (Eq.~\eqref{eq:concat}), revealing a strong bias toward language inputs. The lack of consistent vertical stripes further suggests limited focus on key visual elements, weakening the model’s ability to maintain cross-modal grounding over time. Overall, attention in the baseline appears noisy and scattered across layers, indicating difficulty in forming stable associations between visual content and language queries.

These differences highlight VisAlign’s effectiveness in improving the model’s ability to integrate visual and textual modalities. By promoting more balanced attention, VisAlign improves focus on critical visual cues often overlooked by baseline Video-LLaVA, strengthening temporal and spatial coherence across the transformer layers and boosting overall visual information use.
 
\textit{\textbf{Why VisAlign encourages the model to use visual information?}} VisAlign introduces no additional visual inputs or training objectives; its contribution is purely representational. By augmenting the LLM’s textual token embeddings with averaged visual embeddings, VisAlign alters the model’s input representation and reshapes the optimization landscape, reducing the reliance on memorized textual priors.  As discussed earlier, LVLMs tend to over-attend to text because the underlying LLM is pre-trained exclusively on textual embeddings. During multi-modal training, language tokens remain in-distribution for the backbone LLM, whereas visual tokens are comparatively out-of-distribution, leading to the strong attention bias toward text as observed in Fig. \ref{fig:attention_layers} (top). By injecting visual information into every textual embedding, VisAlign deliberately departs from the pre-training distribution, encouraging the model to adapt its early-layer attention patterns during finetuning and to more consistently incorporate visual evidence.

% By modifying the initial embeddings conditioned by the LLM, it reshapes the optimization landscape and reduces reliance on memorized textual priors. As discussed LVLMs over-attend to text because the LLM is pre-trained exclusively on textual embeddings. During VLM training, language tokens remain familiar to the LLM, whereas visual tokens are not, leading to the strong attention bias observed in Fig. \ref{fig:attention_layers} (top). By injecting visual information into every text embedding, VisAlign systematically departs from the pre-training distribution, encouraging the LLM to adjust its early-layer attention patterns during finetuning.

% the LLM is no longer able to treat text as the primary modality;
% attention becomes more uniformly distributed across modalities, reducing reliance on linguistic shortcuts.
% Collectively, this demonstrates that integrating visual context within the textual embeddings to modify the textual embedding does encourage the model to use visual information more.

\vspace{-2mm}
\section{Experiments and Results} 
\vspace{-2mm}

% This section evaluates VisAlign's effectiveness in reducing hallucinations by comparing it with the baseline across several benchmarks (a detailed description of these benchmarks is provided in Appendix Sec.\ref{appx:dataset}.

We evaluate VisAlign across a diverse set of benchmarks that probe hallucinations and visual grounding from complementary perspectives, including fine-grained visual discrimination, object-level hallucinations, factual consistency under visual edits, sequential visual reasoning, and conflicts between visual input and parametric memory. The results and discussions for each benchmark are presented below (refer Sec.\ref{appx:dataset} for detailed description).
%: of these benchmarks is provided in Appendix Sec.\ref{appx:dataset}):

\begin{figure*}%[h]
   
    \centering    \includegraphics[width=1\textwidth]{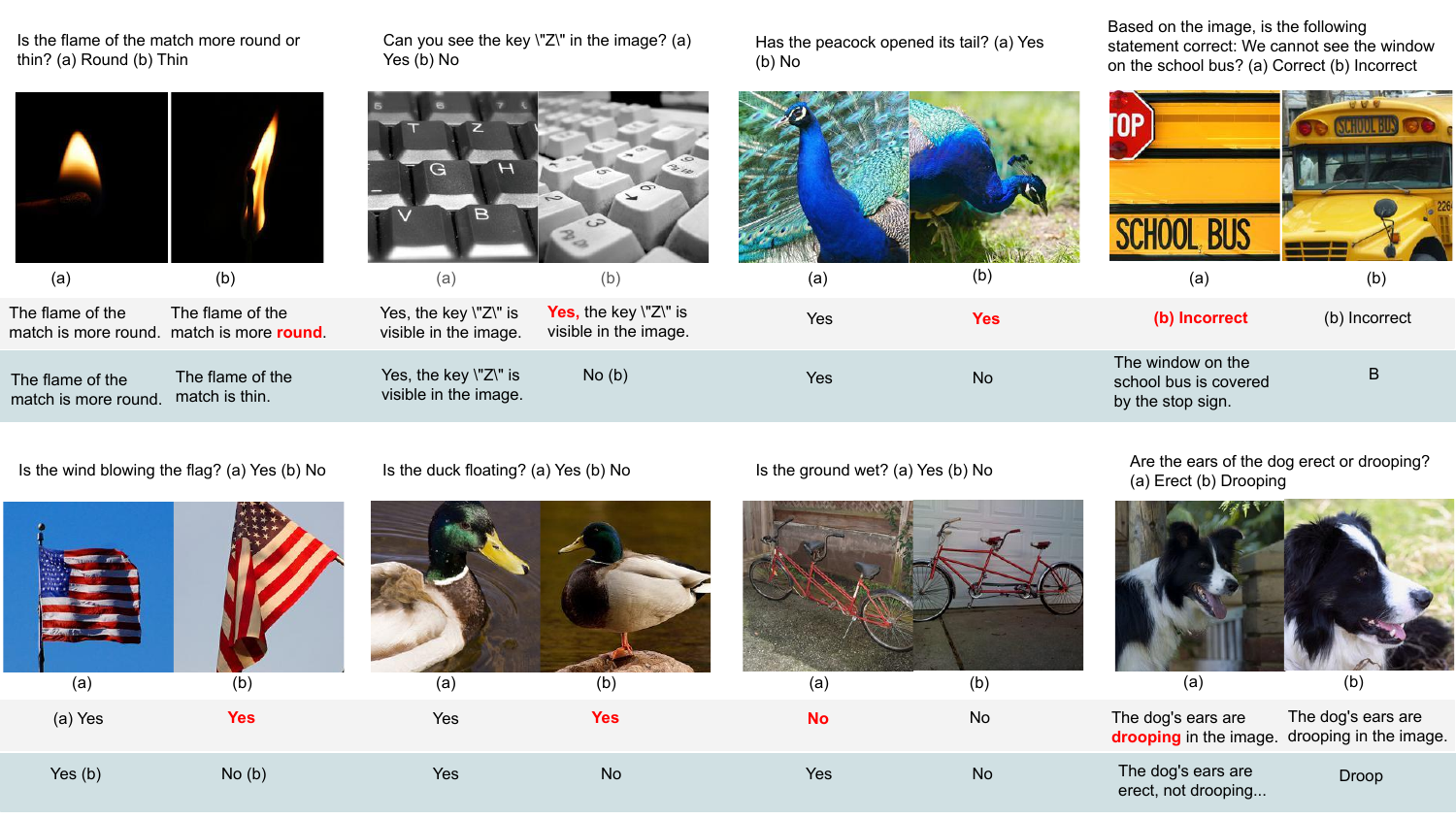}
    % \vspace{-1em}
    \caption{Qualitative results from the \textbf{MMVP-MLLM Benchmark}:
    Below each image, the baseline model's response is shown first, followed by the response from the model trained with \textit{VisAlign}.}
    % \vspace{-2em}
    \label{fig:mmvp_images}
\end{figure*}

\textbf{MMVP-MLLM} The results in Table~\ref{tab:pope_mmvp} show that Video-LLaVA enhanced with VisAlign achieves a substantial \textbf{+9.33\%} improvement over the baseline. Since MMVP-MLLM is specifically designed to probe bias and hallucination in LVLMs by enforcing fine-grained visual discrimination under minimal semantic variance, this gain is especially significant. It demonstrates that VisAlign markedly strengthens the model’s grounding in visual evidence rather than relying on linguistic priors, effectively reducing hallucinations and improving factual consistency. A qualitative comparison is presented in Figure~\ref{fig:mmvp_images}. In the first example, the model must distinguish between two flame images---one round and the other elongated. The baseline Video-LLaVA incorrectly classifies both as “round,” indicating over-reliance on memorized language patterns. In contrast, the VisAlign-enhanced model correctly differentiates the shapes, demonstrating stronger visual grounding. Similar improvements appear in other examples, underscoring VisAlign’s effectiveness in reducing hallucinations and promoting accurate, cross-modal reasoning.

% \begin{table}%[h]
% \small
% \centering
% \resizebox{0.45\textwidth}{!}{\begin{tabular}{lllll|l|l}
% \hline
%                        & \multicolumn{4}{c|}{POPE A-OKVQA}                                  & \multicolumn{2}{l}{MMVP-MLLM} \\ \hline
%                        & \textbf{Acc}       & \textbf{P} & \textbf{R} & \multicolumn{1}{l|}{\textbf{F1}} & \multicolumn{2}{c}{\textbf{Acc}} 
%                        \\
%                        \hline
% Video-LLaVA & 54.1           & 52.14     & 99.6   & \multicolumn{1}{l|}{68.45}    & \multicolumn{2}{c}{14}                 \\
% \ + VisAlign  & \textbf{57.09} & \textbf{53.9}      & 98.33  & \multicolumn{1}{l|}{\textbf{69.63}}    & \multicolumn{2}{c}{\textbf{23.33}}     \\ \hline    
% \end{tabular}}
% \caption{Results on POPE A-OKVQA \citep{li2023evaluating} and MMVP-MLLM \citep{tong2024eyes}. Acc: Accuracy, P:Precision, R:Recall, F1: F1 score.
% % \vspace{-0.5cm}
% }
% \label{tab:pope_mmvp}
% \end{table}

% changing the order of the dataset to match with the order in text

\begin{table}%[h]
\centering
\resizebox{0.45\textwidth}{!}{\begin{tabular}{l|l|lllll}
\hline
                        & \multicolumn{2}{|l|}{MMVP-MLLM} & \multicolumn{4}{c}{POPE A-OKVQA}                                   \\ 
                       \hline
                       
                       & \multicolumn{2}{c|}{\textbf{Acc}} & \textbf{Acc}       & \textbf{P} & \textbf{R} & \multicolumn{1}{l}{\textbf{F1}} 
                       \\
                       \hline
                       
Video-LLaVA & \multicolumn{2}{c|}{14} &  54.1  & 52.14     & 99.6   & \multicolumn{1}{l}{68.45}   
\\
\ + VisAlign  & \multicolumn{2}{c|}{\textbf{23.33}} &  \textbf{57.09} & \textbf{53.9}      & 98.33  & \multicolumn{1}{l}{\textbf{69.63}}         \\ \hline    
\end{tabular}}
\caption{Results on POPE A-OKVQA \citep{li2023evaluating} \& MMVP-MLLM \citep{tong2024eyes}. Acc: Accuracy, P:Precision, R:Recall, F1: F1 score.
% \vspace{-0.5cm}
}
\label{tab:pope_mmvp}
\end{table}

% \begin{wraptable}{r}{0.52\textwidth}
%     \vspace{-1.6em}
%     \centering
%     \footnotesize
%     \setlength{\tabcolsep}{3.5pt}
%     \renewcommand{\arraystretch}{1.05}
%     \resizebox{0.50\textwidth}{!}{%
%     \begin{tabular}{l|l|lllll}
%         \hline
%             & \multicolumn{2}{|l|}{MMVP-MLLM} & \multicolumn{4}{c}{POPE A-OKVQA} \\ 
%         \hline
%             & \multicolumn{2}{c|}{\textbf{Acc}} & \textbf{Acc} & \textbf{P} & \textbf{R} & \textbf{F1} \\
%         \hline
%         Video-LLaVA & \multicolumn{2}{c|}{14.0} & 54.1 & 52.14 & 99.6 & 68.45 \\
%         + VisAlign  & \multicolumn{2}{c|}{\textbf{23.33}} & \textbf{57.09} & \textbf{53.9} & 98.33 & \textbf{69.63} \\
%         \hline
%     \end{tabular}}
%     \vspace{-0.8em}
%     \caption{Results on POPE A-OKVQA \citep{li2023evaluating} and MMVP-MLLM \citep{tong2024eyes}. Acc: accuracy; P: precision; R: recall.}
%     \label{tab:pope_mmvp}
%     \vspace{-1em}
% \end{wraptable}

\begin{figure*}[h]
    \centering
\includegraphics[width=1\textwidth, trim=0 72 0 0, clip]{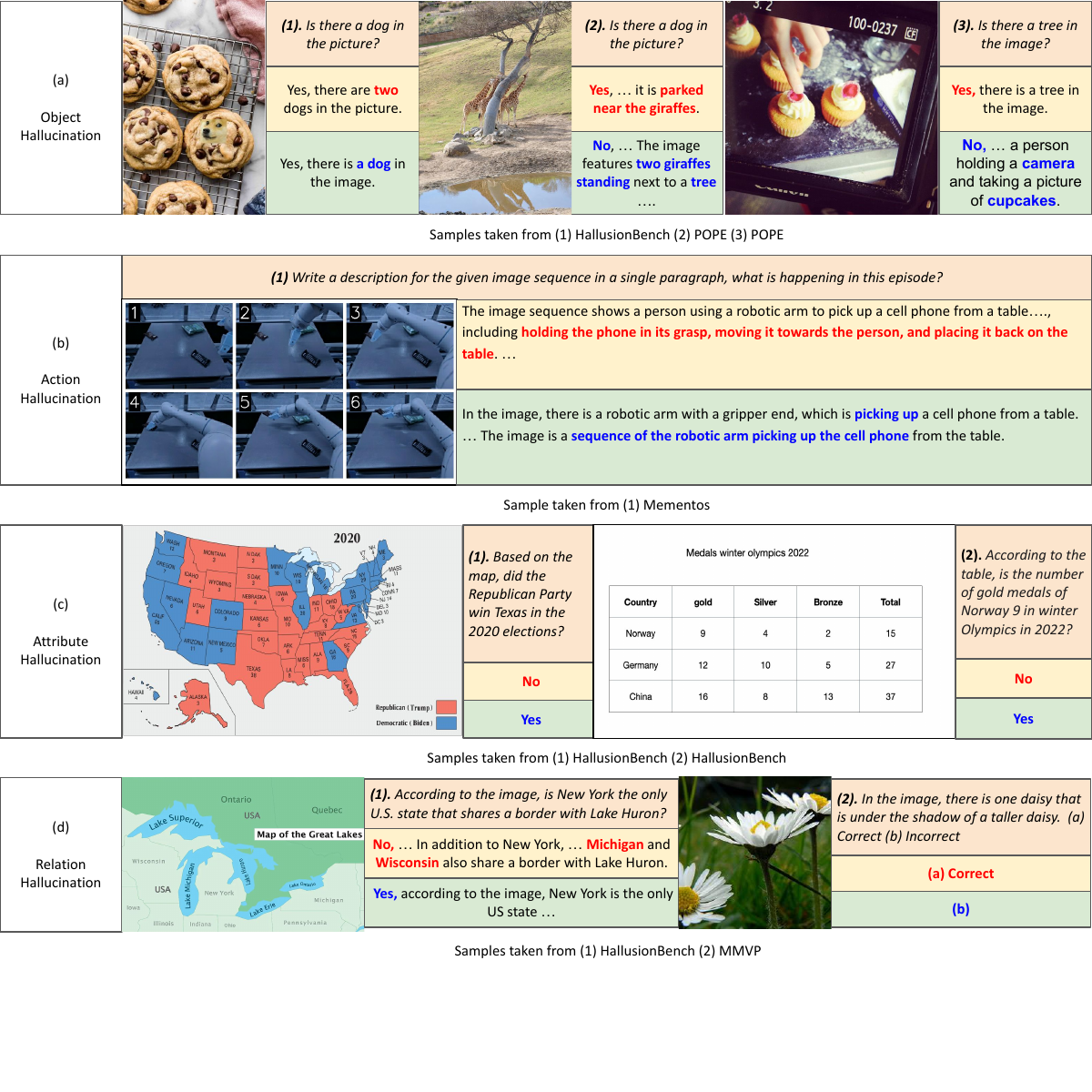}
    % \vspace{-1em}
    \caption{Qualitative examples from \textbf{POPE A-OKVQA, HallusionBench, MMVP}, and \textbf{Mementos} benchmarks illustrating various hallucination types. Input prompts are shown in orange, baseline Video-LLaVA outputs in yellow, and VisAlign-enhanced outputs in green. VisAlign consistently improves performance across object, action, attribute, and relation hallucinations.}
    % \vspace{-2em}
    \label{fig:allbench_images}
\end{figure*}

% \textbf{POPE} Following prior work~\citep{villa2025eagle}, we focus on the most challenging setting: Adversarial SEEM from A-OKVQA, which applies SEEM-based object detection to A-OKVQA images. This subset probes whether models falsely affirm the presence of common yet incorrect objects, revealing object-level hallucinations driven by language bias. Table~\ref{tab:pope_mmvp} presents quantitative results on the POPE benchmark, where VisAlign consistently surpasses the baseline across key metrics, achieving a \textbf{2.99\%} increase in accuracy, a \textbf{1.76\%} boost in precision, and a \textbf{1.18\%} gain in F1-score. The notable rise in precision indicates a significant reduction in false positives---hallucinated objects---while the improved F1-score reflects a more robust balance between precision and recall. These provide strong evidence that VisAlign effectively curtails predictions of frequent yet visually unsupported objects, thereby substantially enhancing object-level visual grounding. Supporting qualitative results in Fig.~\ref{fig:allbench_images} further reinforce VisAlign’s reliability in avoiding erroneous affirmations of absent objects, underscoring its role in advancing cross-modal integration and reducing hallucinations.

%slightly revised for ACL
\textbf{POPE} Following prior work~\citep{villa2025eagle}, we evaluate on the most challenging setting: Adversarial SEEM from A-OKVQA, which applies SEEM-based object detection to A-OKVQA images. This subset probes whether models incorrectly affirm the presence of common but absent objects, revealing object-level hallucinations driven by language bias. As shown in Table~\ref{tab:pope_mmvp}, VisAlign consistently outperforms the baseline on the POPE benchmark, improving accuracy by \textbf{2.99\%}, precision by \textbf{1.76\%}, and F1-score by \textbf{1.18\%}. The gain in precision indicates fewer false positives (hallucinated objects), while the improved F1 reflects a better precision–recall balance. Qualitative examples in Fig.~\ref{fig:allbench_images} further illustrate VisAlign’s ability to avoid affirming visually unsupported objects, highlighting its effectiveness in improving visual grounding.

\textbf{MERLIN} evaluates factual consistency and visual grounding in LVLMs through fine-grained object existence verification. Table~\ref{tab:merlin} presents quantitative results for both positive (object present) and negative (object removed) cases, evaluated under two distinct image sampling strategies. VisAlign consistently outperforms the baseline demonstrating superior capability to mitigate hallucinations by enhancing the model’s sensitivity to subtle visual cues, thereby substantially improving visual fidelity and robustness in fine-grained, object-centric reasoning tasks.

\begin{table}[]
\centering
\resizebox{0.48\textwidth}{!}{%
\footnotesize
\begin{tabular}{lllll}
\hline
            & \multicolumn{4}{c}{\textbf{Curated Images}}                                \\ \hline
            & Pos-Orig & Pos-Edited & Neg-Orig & Neg-Edited \\ \hline
Video-LLava & 30.9          & 16.7            & 71.5          & 79.6            \\
VisAlign    & \textbf{34.3} & \textbf{20.3}   & \textbf{72.7} & \textbf{83.0}   \\ \hline
            & \multicolumn{4}{c}{\textbf{Random Images}}                                 \\ \hline
            & Pos-Orig & Pos-Edited & Neg-Orig & Neg-Edited \\ \hline
Video-LLava & 48.2          & 33.3            & 59.5          & 67.9            \\
VisAlign    & \textbf{48.6} & \textbf{36.7}   & \textbf{60.1} & \textbf{71.3}   \\ \hline
\end{tabular}%
}
\caption{Results (in \%) on the Merlin benchamark \cite{villa2023behind}). ``Pos":Positive, ``Neg":Negative.}
\vspace{-1mm}
\label{tab:merlin}
\end{table}

% \begin{wraptable}{r}{0.50\textwidth}
%     \vspace{-0.8em}
%     \centering
%     \footnotesize
%     \setlength{\tabcolsep}{4pt}
%     \renewcommand{\arraystretch}{1.05}
%     \resizebox{0.48\textwidth}{!}{%
%     \begin{tabular}{lllll}
%         \hline
%             & \multicolumn{4}{c}{\textbf{Curated Images}} \\ \hline
%             & Pos-Orig & Pos-Edited & Neg-Orig & Neg-Edited \\ \hline
%         Video-LLava & 30.9 & 16.7 & 71.5 & 79.6 \\
%         VisAlign    & \textbf{34.3} & \textbf{20.3} & \textbf{72.7} & \textbf{83.0} \\ \hline
%             & \multicolumn{4}{c}{\textbf{Random Images}} \\ \hline
%             & Pos-Orig & Pos-Edited & Neg-Orig & Neg-Edited \\ \hline
%         Video-LLava & 48.2 & 33.3 & 59.5 & 67.9 \\
%         VisAlign    & \textbf{48.6} & \textbf{36.7} & \textbf{60.1} & \textbf{71.3} \\ \hline
%     \end{tabular}}
%     \vspace{-0.8em}
%     \caption{Results (\%) on the Merlin benchmark \cite{villa2023behind}. ``Pos'': positive, ``Neg'': negative.}
%     \label{tab:merlin}
%     \vspace{-1.5em}
% \end{wraptable}

\textbf{Mementos} evaluates sequential image reasoning in LVLMs across three domains: \textit{Robotics}, \textit{Comics}, and \textit{Daily Life}. It rigorously test object and action hallucinations within dynamic visual contexts, emphasizing temporal coherence and object-behavior relationships. This makes Mementos especially valuable for assessing a multimodal model’s ability to detect hallucinations while accurately understanding complex, evolving visual narratives.

Table~\ref{tab:mementos} shows significant improvements in the \textit{Robotics} domain for both object hallucination (+\textbf{1.13}\% accuracy) and action hallucination (+\textbf{0.97}\% accuracy). These gains stem from the structured, goal-driven nature of robotic sequences, where predictable temporal patterns and clear visual cues enable VisAlign to maintain coherent attention over time and better align visual tokens with text, enhancing temporal reasoning of object states and behaviors. In contrast, improvements in the \textit{Comics} and \textit{Daily Life} domains are more modest, likely due to their greater visual and semantic complexity. Comics often use stylized, symbolic imagery and abstract narratives that disrupt typical visual-linguistic links, while Daily Life scenes involve high variability, subtle object transitions, and complex human actions that hinder consistent temporal alignment. In these unstructured contexts, VisAlign’s attention calibration is limited by noisier, less reliable visual inputs.

% \textbf{HallusionBench} \citep{guan2024hallusionbench} We provide results in Appendix Section \ref{appx:HallusionBench}.

% updated table for saving space

\begin{table}[t]
\centering
\scriptsize
\setlength{\tabcolsep}{3pt}
\begin{tabular}{lcccccccc}
\hline
 & \multicolumn{4}{c}{\textbf{Object}} & \multicolumn{4}{c}{\textbf{Action}} \\
\cmidrule(lr){2-5} \cmidrule(lr){6-9}
\textbf{Method} & Acc. & Prec. & Rec. & F1 & Acc. & Prec. & Rec. & F1 \\
\midrule
\multicolumn{9}{c}{\textit{Robotics Domain}} \\
Video-LLaVA & 8.27 & 16.55 & 12.40 & 13.46 & 5.53 & 6.99 & \textbf{11.30} & 8.40 \\
+VisAlign   & \textbf{9.40} & \textbf{19.20} & \textbf{13.61} & \textbf{15.16} & \textbf{6.50} & \textbf{9.60} & 10.76 & \textbf{9.45} \\
\midrule
\multicolumn{9}{c}{\textit{Daily Life Domain}} \\
Video-LLaVA & 22.05 & 38.30 & 31.90 & 33.55 & 13.50 & 31.70 & \textbf{18.66} & \textbf{22.40} \\
+VisAlign   & \textbf{22.18} & \textbf{38.31} & \textbf{32.31} & \textbf{33.70} & 12.31 & \textbf{32.10} & 16.44 & 20.70 \\
\midrule
\multicolumn{9}{c}{\textit{Comics Domain}} \\
Video-LLaVA & 11.12 & 21.00 & \textbf{19.00} & \textbf{18.86} & \textbf{4.48} & 11.28 & \textbf{6.58} & \textbf{8.08} \\
+VisAlign   & \textbf{12.00} & 21.00 & 17.80 & 18.41 & 4.00 & \textbf{13.33} & 5.36 & 7.10 \\
% \bottomrule
\hline
\end{tabular}
\caption{Results on Mementos~\citep{wang2024mementos} across object and action hallucinations in three domains.}
\label{tab:mementos}
% \vspace{-0.7cm}
\end{table}

\textbf{HallusionBench}
evaluates hallucinations caused by conflicts between visual input and parametric memory in LVLMs, distinguishing Visual-Dependent (VD) and Visual-Supplement (VS) questions and including a hard split with human-edited images that induce modality conflicts. 
On this benchmark, the VisAlign achieves an average improvement of approximately \textbf{3\%}, with consistent gains across both VD and VS categories (detailed results in Appendix \ref{appx:HallusionBench}). The improvements are particularly notable in VS tasks such as Map, OCR, and Table, where reliance on memorized knowledge is most error-prone, indicating that VisAlign effectively encourages visual grounding over language priors. Qualitative examples (Figure~\ref{fig:allbench_images}) further show that VisAlign enables correct interpretation of manipulated visual inputs---such as falsified maps and statistics---where the baseline model hallucinates. Overall, these results demonstrate that VisAlign consistently improves visual grounding and reduces hallucinations under adversarial visual–textual discrepancies.

% \textbf{Summary:} consistent improvements across all benchmarks demonstrate that refining attention score distributions effectively reduces hallucinations, enabling predictions grounded in visual evidence rather than memorized associations.  

% added for ACL
\textbf{Summary:} The consistent improvements across hallucination benchmarks demonstrate that refining attention score distributions helps reduce hallucinations by encouraging predictions grounded in visual evidence rather than memorized associations. To assess the broader impact of VisAlign, we also evaluate on the MME benchmark (Appendix~\ref{general lvlm}). The results show improvements in visually grounded categories (e.g., Existence, Count, Color) and slight declines in knowledge-heavy categories (e.g., Celebrity, Landmark), suggesting stronger reliance on visual evidence.

\begin{table}%[t]
\centering
\small
\resizebox{0.45\textwidth}{!}{\begin{tabular}{lcccc}
\toprule
\textbf{Model} & \textbf{Acc} & \textbf{Precision} & \textbf{Recall} & \textbf{F1-Score} \\
\midrule
LLaVA1.5  (\%)     & 69  & 62.23 & 97.66 & 76.02 \\
+ VisAlign    (\%)     & \textbf{71} & \textbf{64}  & 97.13 & \textbf{77.01} \\
\midrule
OpenQwen2VL  (\%)     & 53.13  & 80.12 & 8.33 & 15.09 \\
+ VisAlign    (\%)     & \textbf{55.7} & 56.8  & \textbf{47.6} & \textbf{51.8} \\
\bottomrule
\end{tabular}}
\caption{Effects of VisAlign on the LLaVA1.5 and OpenQwen2VL baseline, on the POPE-AOKVQA benchmark.}
\label{tab:llava}
\vspace{-3mm}
\end{table}

% \begin{wraptable}{r}{0.46\textwidth}
%     \vspace{-1.4em}
%     \centering
%     \footnotesize
%     \setlength{\tabcolsep}{4pt}
%     \renewcommand{\arraystretch}{1.05}
%     \resizebox{0.46\textwidth}{!}{
%     \begin{tabular}{lcccc}
%         \toprule
%         \textbf{Model} & \textbf{Acc} & \textbf{Prec.} & \textbf{Recall} & \textbf{F1} \\
%         \midrule
%         LLaVA1.5 (\%) & 69.0 & 62.23 & 97.66 & 76.02 \\
%         + VisAlign (\%) & \textbf{71.0} & \textbf{64.0} & 97.13 & \textbf{77.01} \\
%         \midrule
%         OpenQwen2VL (\%) & 53.13 & 80.12 & 8.33 & 15.09 \\
%         + VisAlign (\%) & \textbf{55.7} & 56.8 & \textbf{47.6} & \textbf{51.8} \\
%         \bottomrule
%     \end{tabular}}
%     \vspace{-0.8em}
%     \caption{Effect of VisAlign on LLaVA1.5 and OpenQwen2VL on POPE-AOKVQA.}
%         \vspace{-1.3em}
%     \label{tab:llava}

% \end{wraptable}

\textbf{Performance on Additional LVLMs:}
To further validate the generality and robustness of VisAlign, we evaluate its effectiveness on two SOTA LVLM, LLaVA 1.5 and Open-Qwen2VL. As shown in Table~\ref{tab:llava}, VisAlign consistently enhances performance and reduces hallucinations highlighting its broad applicability and effectiveness across different LVLMs (further details: refer Sec. \ref{appx:models}).

% \textbf{Qualitative Analysis:} We show examples from various benchmarks in Figs.~\ref{fig:mmvp_images} and \ref{fig:allbench_images} (additional visualizations in the appendix). These examples illustrate how VisAlign reduces hallucinations compared to baseline LVLMs. For instance, in Fig.~\ref{fig:allbench_images}, the first image shows cookies with a dog photo affixed on top; the question “Is there a dog in the picture?” is incorrectly answered as “two dogs” by the baseline, while VisAlign correctly predicts “one dog.” In the second example, the baseline predicts dogs on an image of two giraffes, whereas VisAlign gives the correct answer. Similar improvements are observed across other cases, demonstrating that VisAlign enhances visual grounding by incorporating visual context into textual embeddings.

\begin{table}%[ht]
\centering
\small
\resizebox{0.45\textwidth}{!}{\begin{tabular}{lcccc}
\toprule
\textbf{Model} & \textbf{Acc} & \textbf{Precision} & \textbf{Recall} & \textbf{F1-Score} \\
\midrule
Video-LLaVA       & 54.1  & 52.14 & \textbf{99.6} & 68.45 \\
+ VCD~\citep{leng2024mitigating}              & 54.5  & 52.38 & 99.39 & 68.6 \\
+ VisAlign         & 57.09 & 53.9  & 98.33 & 69.63 \\
\textbf{+ VisAlign + VCD} & \textbf{58.8}  & \textbf{55.03} & 96.33 & \textbf{70.04} \\
\bottomrule
\end{tabular}}
\caption{Performance comparison of Video-LLava with existing hallucination mitigation approaches on POPE-AOKVQA. 
}
\label{tab:pope_vcd}
% \vspace{-5mm}
\end{table}
\textbf{Comparison with Existing Hallucination Mitigation Approaches:}
This section compares VisAlign with other SOTA hallucination mitigation methods. 
We focus on Visual Contrastive Decoding (VCD)~\citep{leng2024mitigating}, a strong inference-time SOTA method. While model-agnostic and lightweight, such inference-time methods complement VisAlign, which proactively mitigates hallucinations by refining representations during training (refer Sec. \ref{appx:VCD} in Appendix for further details on VCD). Table~\ref{tab:pope_vcd} compares the effectiveness of VisAlign and VCD applied to Video-LLaVA, both individually and combined. VisAlign outperforms VCD alone, notably improving accuracy (\textbf{54.5 → 57.09}) and F1-score (\textbf{68.6 → 69.63}). While VCD delivers incremental gains by refining output selection during inference, VisAlign achieves more substantial improvements by addressing modality imbalance during training. When combined, the two methods yield the best overall performance, further boosting accuracy to \textbf{58.8} and F1-score to \textbf{70.04}, demonstrating their complementary strengths.

% \begin{wraptable}{r}{0.46\textwidth}
%     \vspace{-1.5em}
%     \centering
%     \footnotesize
%     \setlength{\tabcolsep}{4pt}
%     \renewcommand{\arraystretch}{1.05}
%     \resizebox{0.46\textwidth}{!}{
%     \begin{tabular}{lcccc}
%         \toprule
%         \textbf{Model} & \textbf{Acc} & \textbf{Prec.} & \textbf{Recall} & \textbf{F1} \\
%         \midrule
%         Video-LLaVA & 54.1 & 52.14 & \textbf{99.6} & 68.45 \\
%         + VCD~\citep{leng2024mitigating} & 54.5 & 52.38 & 99.39 & 68.60 \\
%         + VisAlign & 57.09 & 53.90 & 98.33 & 69.63 \\
%         \textbf{+ VisAlign + VCD} & \textbf{58.8} & \textbf{55.03} & 96.33 & \textbf{70.04} \\
%         \bottomrule
%     \end{tabular}}
%     \vspace{-0.8em}
%     \caption{Performance of Video-LLaVA with hallucination mitigation methods on POPE-AOKVQA.}
%     % \vspace{-1em}
%     \label{tab:pope_vcd}  
% \end{wraptable}

These results underscore VisAlign’s orthogonality to inference-time techniques like VCD, allowing it to enhance performance without interference. Also highlighting its strong generalizability---VisAlign’s benefits persist even when integrated with other hallucination mitigation strategies, showcasing its robustness across diverse settings.

% \vspace{-2mm}
\section{Conclusion} 
% \vspace{-2mm}
% We systematically analyze attention distributions in LVLMs concerning hallucinations---outputs lacking grounding in visual input. Our findings show that popular LVLMs overemphasize text over visual information, increasing reliance on linguistic priors and hallucinations. To address this, we propose a simple yet effective method that redefines textual embeddings, rebalancing attention during training and improving the model’s use of visual information. This results in significantly reduced hallucinations and more semantically accurate, visually faithful outputs.
% We validate our approach across multiple challenging hallucination benchmarks, consistently achieving substantial improvements. We hope these insights inspire further research to better leverage visual data, reduce hallucinations, and enhance the reliability of multimodal reasoning in LVLMs.

We systematically analyze attention distributions in LVLMs with respect to hallucinations---outputs that lack grounding in visual input---and find that popular models overemphasize text, leading to increased reliance on linguistic priors. To address this, we propose a simple yet effective method that redefines textual embeddings to rebalance attention during training and improve the use of visual information. This results in significantly reduced hallucinations and more semantically accurate, visually grounded outputs. We validate our approach across multiple challenging hallucination benchmarks, consistently achieving substantial improvements. We hope these findings inspire further research toward better leveraging visual data and enhancing the reliability of multimodal reasoning in LVLMs.

% \clearpage
\section{Limitations}
% added for ACL
In this work, we address modality imbalance in LVLMs by modifying multimodal interaction at the representation level, aiming to improve visual grounding and reduce hallucinations. To maintain a lightweight and architecture-agnostic design, VisAlign integrates visual information into textual embeddings using globally average-pooled visual features. While this global aggregation allows us to isolate the effect of representation-level alignment, more sophisticated pooling strategies—such as attention-based aggregation over region-level visual tokens—could capture finer-grained visual cues and further strengthen cross-modal alignment, potentially leading to additional performance gains.
Because VisAlign encourages stronger reliance on visual evidence, tasks that depend more heavily on memorized world knowledge or linguistic priors may experience slight performance declines. Our analysis on the MME benchmark shows minor drops in categories such as Landmark and Celebrity, suggesting a potential trade-off between visual grounding and language-driven reasoning.
Finally, although we evaluate VisAlign on several hallucination benchmarks, broader validation on more recent LVLM architectures and large-scale multimodal benchmarks would provide further insight into its general applicability.

\clearpage

\section{Acknowledgment}
Agrawal and Huang are supported by DARPA Transfer from Imprecise and Abstract Models to Autonomous Technologies (TIAMAT) 80321, DARPA HR001124S0029-AIQ-FP-019, DOD-AFOSR-Air Force Office of Scientific Research under award number FA9550-23-1-0048, National Science Foundation TRAILS Institute (2229885). Private support was provided by Peraton and Open Philanthropy. The Authors acknowledge the National Artificial Intelligence Research Resource (NAIRR) Pilot and [insert the resources supporting your project here] for contributing to this research result.

\bibliography{custom}
\clearpage
\appendix
\section{Appendix}
\label{appendix}

\subsection{Related Works}

\textbf{Hallucination Detection and Mitigation in LVLMs} is actively studied through a range of benchmarks designed to evaluate diverse hallucination types. POPE-AOKVQA~\citep{li2023evaluating} and NOPE~\citep{lovenia2023nope} focus on object-level hallucinations, while MERLIN~\citep{jing2023faith} examines factual consistency via atomic fact decomposition. MMVP-MLLM~\citep{tong2024eyes} and HallusionBench~\citep{guan2024hallusionbench} probe model behavior under minimal semantic variation and cross-modal conflicts. Mementos~\citep{wang2024mementos} targets temporal hallucinations in sequential visual reasoning. AMBER~\citep{wang2023amber} introduces a unified benchmark for evaluating both discriminative and generative hallucinations. Together, these datasets reveal a broad spectrum of hallucination phenomena---including object, action, attribute, relational, and temporal inconsistencies---highlighting the complexity of achieving reliable visual grounding in LVLMs.

\subsection{Detailed Description of Various Benchamarks used in Evaluating our method} \label{appx:dataset}

\textbf{MMVP-MLLM}~\citep{tong2024eyes} benchmark features carefully curated image pairs with highly similar CLIP embeddings, minimizing semantic divergence and emphasizing subtle visual distinctions. Each pair is accompanied by two binary-choice questions targeting fine-grained visual understanding. A model receives credit only if it answers both correctly, enforcing a strict criterion that rewards accurate visual grounding and penalizes reliance on language priors. This makes MMVP-MLLM particularly effective for evaluating hallucinations, as it compels models to rely on actual visual evidence rather than linguistic shortcuts or memorized associations.

\textbf{POPE}~\citep{li2023evaluating} evaluates hallucinations through yes/no questions about object presence in images. “Yes” questions correspond to ground-truth objects, while “No” questions are adversarially crafted from the top-k most frequent object categories absent from the image. This setup exposes the model’s reliance on language priors by testing its ability to reject visually unsupported but common objects. Following prior work~\citep{villa2025eagle}, we focus on the most challenging setting: Adversarial SEEM from A-OKVQA, which applies SEEM-based object detection to A-OKVQA images. This subset probes whether models falsely affirm the presence of common yet incorrect objects, revealing object-level hallucinations driven by language bias. POPE thus offers a fine-grained, targeted measure of visual grounding, serving as a rigorous and complementary benchmark to evaluate VisAlign’s effectiveness in reducing hallucinations.

\textbf{MERLIN}~\cite{villa2023behind} evaluates factual consistency and visual grounding in LVLMs through fine-grained object existence verification. It employs a curated set of original and synthetically edited images to assess whether models can accurately detect the presence or absence of objects. Our evaluation specifically targets a subset of MERLIN where an entire object category, limited to a single instance in the original image, has been removed in the edited version.

\textbf{Mementos}~\citep{wang2024mementos} evaluates sequential image reasoning in LVLMs across three domains: \textit{Robotics}, \textit{Comics}, and \textit{Daily Life}. It rigorously test object and action hallucinations within dynamic visual contexts, emphasizing temporal coherence and object-behavior relationships. This makes Mementos especially valuable for assessing a multimodal model’s ability to detect hallucinations while accurately understanding complex, evolving visual narratives.

\textbf{HallusionBench} \citep{guan2024hallusionbench} is a diagnostic benchmark assessing how parametric memory affects hallucinations in LVLMs. It categorizes questions into Visual-Dependent (VD), requiring visual input, and Visual-Supplement (VS), answerable using world knowledge or training data. VS questions evaluate the model’s ability to resolve conflicts between visual input and parametric memory. The benchmark includes easy and hard splits, with the hard subset featuring human-edited images designed to create modality conflicts.

\subsection{Results on HallusionBench Benchmark}\label{appx:HallusionBench}

\begin{table*}%[htp]
\centering
\resizebox{0.83\textwidth}{!}{%
\begin{tabular}{lllllllllll}
                                                         
\hline 
                      
                                        & \multicolumn{5}{l}{Visual Dependent} & \multicolumn{4}{l}{Visual Supplement} &  \\
                                        \hline
Method                 & Figure & Ilusion & Math  & OCR   & Video & Chart & Map   & OCR   & Table & Average \\
\hline
 & \multicolumn{10}{c}{Hard Data Split}                                               \\ \hline
Video-LLaVA  & 29.27  & 54.93   & 35.29 & 41.30 & 36.84 & 24.56 & 25.00 & 18.52 & 28.79 &     32.72    \\
Video-LLaVA + VisAlign & \textbf{34.15} & 49.30 & \textbf{37.25} & \textbf{45.65} & \textbf{36.84} & 21.05 & \textbf{28.12} & \textbf{33.33} & \textbf{34.85} & \textbf{35.61} \\ \hline
                       & \multicolumn{10}{c}{Easy Data Split}                                               \\ \hline
Video-LLava & 64.10  & 40.28   & 27.78 & 75.61 & 15.94 & 35.11 & 46.88 & 53.70 & 36.36 &   \textbf{43.97}      \\
Video-LLava + VisAlign & 53.85          & 36.11 & \textbf{37.04} & 53.66          & \textbf{36.23} & 25.95 & 48.44          & 50.00          & 28.57          &  41.1
\\
\hline
\end{tabular}%
}
\caption{Category-wise results on the HallusionBench benchmark\citep{guan2024hallusionbench}.}
\label{tab:HallusionBench}
\vspace{-0.5cm}
\end{table*}

\textbf{HallusionBench} \citep{guan2024hallusionbench} is a diagnostic benchmark assessing how parametric memory affects hallucinations in LVLMs. It categorizes questions into Visual-Dependent (VD), requiring visual input, and Visual-Supplement (VS), answerable using world knowledge or training data. VS questions evaluate the model’s ability to resolve conflicts between visual input and parametric memory. The benchmark includes easy and hard splits, with the hard subset featuring human-edited images designed to create modality conflicts.

Table~\ref{tab:HallusionBench} shows an average improvement of about \textbf{3\%} on the challenging hard subset. Significant gains are seen in Visual-Dependent (VD) tasks, with improvements of \textbf{4.88\%}, \textbf{1.96\%}, and \textbf{4.35\%} in the “Figure,” “Math,” and “OCR” categories, respectively. Even larger gains occur in Visual-Supplement (VS) tasks, with \textbf{3.12\%}, \textbf{14.81\%}, and \textbf{6.06\%} improvements in “Map,” “OCR,” and “Table.” These results are particularly notable because the hard subset contains human-edited images designed to conflict with common knowledge, forcing the model to rely on visual input rather than memorized facts. The gains indicate that VisAlign substantially improves the model’s ability to ground predictions in visual evidence, reducing over-reliance on language priors. For example, Figure~\ref{fig:allbench_images} (d)(1) shows a manipulated map where New York is falsely depicted bordering Lake Huron; while baseline Video-LLaVA hallucinates based on memorized geography, Video-LLaVA+VisAlign correctly interprets the altered visual context. Similarly, in (c)(2), a falsified medal count for Norway is accurately detected only by the VisAlign-enhanced model. These examples 
highlight VisAlign’s effectiveness in enhancing visual grounding and mitigating hallucinations by improving sensitivity to subtle visual inconsistencies.

\subsection{Effect of VisAlign on MME, a General LVLM benchmark:} \label{general lvlm}
In the main paper, we comprehensively evaluated VisAlign’s effectiveness in reducing hallucinations across multiple benchmarks, consistently demonstrating significant and robust improvements. Although our primary focus is on hallucination tasks, to further investigate VisAlign’s broader impact, we also assess how it influences the baseline model’s performance on generic vision-language understanding benchmarks.

To this end, we evaluate on the MME benchmark~\citep{mme}, a widely adopted diagnostic suite designed to probe the general capabilities of LVLMs. MME includes various subcategories covering fine-grained visual understanding and textual grounding tasks, such as Existence, Count, Position, Color, Posters, Celebrity, Scene, Landmark,  Artwork,  and OCR. These categories span a range of difficulty, from low-level visual perception to high-level semantic reasoning, offering a comprehensive lens into overall model competency.

Table~\ref{tab:vcd_mme} reports category-wise performance comparing the baseline Video-LLaVA, VisAlign and VCD augmented versions.  
VisAlign significantly improves upon the baseline in several key subcategories that are sensitive to visual grounding, such as Existence (\textbf{170 → 190}), Count (\textbf{121.66 → 131.66}), and Color (\textbf{135 → 148.33}). These improvements align with the primary objective of VisAlign---mitigating hallucinations by enhancing the model’s attention to visual evidence---demonstrating its positive influence on tasks that demand precise object recognition and attribute understanding. Moreover, in categories such as OCR and Posters, VisAlign preserves the same level of performance as the baseline, indicating that it does not compromise tasks unrelated to hallucination-prone scenarios. However, some categories---such as Position, Celebrity, Scene, Landmark, and Artwork---show drop in performance. These tasks often require fine-grained spatial reasoning or prior world knowledge, which may be subtly impacted by VisAlign’s architectural shift toward reinforcing visual embeddings over memorized linguistic patterns. This suggests that while VisAlign strengthens core visual grounding, it may introduce minor trade-offs in more specialized or context-dependent tasks.

Another observation from Table~\ref{tab:vcd_mme} is that state-of-the-art hallucination mitigation methods like \textbf{VCD cause a universal performance drop or yield no improvements across all MME subcategories}. In contrast, VisAlign demonstrates a more favorable trade-off: while it introduces minor performance reductions in certain high-level categories, it provides targeted improvements in core grounding tasks without degrading overall reliability. This contrast highlights VisAlign’s orthogonality to inference-time methods and its potential to improve multimodal reasoning in a more integrated and generalizable manner.

In summary, while VisAlign is primarily designed to mitigate hallucinations, it also brings positive side effects on general VLM tasks that benefit from stronger visual grounding. By enriching textual embeddings with visual information, VisAlign promotes faithful grounding in visual inputs and reduces over-reliance on language priors. Unlike inference-time methods like VCD---which often reduce performance on generic benchmarks---VisAlign improves internal representations, preserving or enhancing accuracy in key subcategories like Color, Count, and Existence. However, this stronger grounding can slightly reduce performance in tasks relying on memorized knowledge or abstract reasoning (e.g., Landmark or Celebrity), due to reduced influence from language-driven biases. This trade-off is expected and could potentially be mitigated by training on larger-scale multimodal datasets—an exciting direction for future work. Overall, VisAlign offers a principled, generalizable, and training-efficient approach to hallucination reduction while preserving broader multimodal capabilities.

\subsection{Comparison with existing hallucination mitigation approaches} \label{appx:VCD} In the main paper, we showed that VisAlign significantly reduces hallucinations in Video-LLaVA by improving the attention score distribution across visual and textual modalities. In this section, we extend our analysis by comparing VisAlign with other state-of-the-art (SOTA) hallucination mitigation methods. As noted in the Related Work section (\ref{related}), inference-time strategies currently represent the leading approaches for mitigating hallucinations. These methods intervene during the decoding stage to guide the model toward generating outputs that are more aligned with the visual input.

We focus on Visual Contrastive Decoding (VCD)~\citep{leng2024mitigating}, a strong inference-time SOTA method. VCD introduces a contrastive re-ranking mechanism, wherein multiple candidate responses are sampled from the model and scored based on both linguistic likelihood and visual alignment. This alignment is computed using a cross-modal similarity function that penalizes syntactically fluent yet visually inconsistent outputs. By re-ranking candidates, VCD encourages the model to favor generations that are both semantically coherent and grounded in the visual input---effectively reducing hallucinations without additional fine-tuning. While model-agnostic and lightweight, such inference-time methods complement VisAlign, which proactively mitigates hallucinations by refining representations during training.

Table~\ref{tab:pope_vcd} compares the effectiveness of VisAlign and VCD applied to Video-LLaVA, both individually and combined. VisAlign outperforms VCD alone, notably improving accuracy (\textbf{54.5 → 57.09}) and F1-score (\textbf{68.6 → 69.63}). While VCD delivers incremental gains by refining output selection during inference, VisAlign achieves more substantial improvements by addressing modality imbalance during training. When combined, the two methods yield the best overall performance, further boosting accuracy to \textbf{58.8} and F1-score to \textbf{70.04}, demonstrating their complementary strengths.

These results underscore VisAlign’s orthogonality to inference-time techniques like VCD, allowing it to enhance performance without interference. They also highlight its strong generalizability---VisAlign’s benefits persist even when integrated with other hallucination mitigation strategies, showcasing its robustness across diverse settings.

\begin{table*}[h]
\centering
\resizebox{\linewidth}{!}{%
\begin{tabular}{lccccccccccc}
\toprule
\textbf{Model} & \textbf{Existence} & \textbf{Count} & \textbf{Position} & \textbf{Color} & \textbf{Posters} & \textbf{Celebrity} & \textbf{Scene} & \textbf{Landmark} & \textbf{Artwork} & \textbf{OCR} \\
\midrule
Video-LLaVA & 170 & 121.66 & \textbf{88.33} & 135 & 103.74 & \textbf{101.47} & \textbf{163} & \textbf{161} & \textbf{107} & 87.5 \\
 +VCD~\citep{leng2024mitigating} & 170 & 105.00 & 76.66 & 125 & 100.00 & 100.88 & 155.75 & 154.5 & 99.25 & 77.5 \\
\textbf{+VisAlign} & \textbf{190} & \textbf{131.66} & 53.33 & \textbf{148.33} & 103.06 & 78.24 & 151 & 125 & 94 & 87.5 \\
% \textbf{ + VisAlign + VCD} & 180 & \textbf{138.33} & 58.33 & \textbf{153.33} & 100 & 79.40 & 141.5 & 122 & 95.75 & 70 \\
\bottomrule
\end{tabular}
}
\caption{Comparison of baseline Video-LLava with different combination of hallucination mitigation approaches on \textbf{MME}.}
\label{tab:vcd_mme}
\end{table*}

\subsection{Performance on additional baselines}\label{appx:models}
\subsubsection{LLava1.5}
In the main paper, we demonstrated that VisAlign significantly reduces hallucinations in Video-LLaVA by improving attention distribution. To further validate the generality and robustness of VisAlign, we evaluate its effectiveness on another state-of-the-art LVLM, LLaVA 1.5~\citep{llava1.5}. As shown in Table~\ref{tab:llava}, VisAlign consistently enhances performance and reduces hallucinations when integrated into this baseline as well. These results highlight the broad applicability and effectiveness of the proposed approach across different LVLMs.

% \begin{table}[t]
% \centering
% % \small
% \resizebox{0.45\textwidth}{!}{\begin{tabular}{lcccc}
% \toprule
% \textbf{Model} & \textbf{Acc} & \textbf{Precision} & \textbf{Recall} & \textbf{F1-Score} \\
% \midrule
% LLaVA1.5  (\%)     & 69  & 62.23 & 97.66 & 76.02 \\
% + VisAlign    (\%)     & \textbf{71} & \textbf{64}  & 97.13 & \textbf{77.01} \\
% \bottomrule
% \end{tabular}}
% \caption{Effects of VisAlign on the OpenQwen2VL baseline, on the POPE-AOKVQA benchmark}
% \label{tab:llava}
% \end{table}

\subsubsection{Open-Qwen2VL}
To further validate the generality and robustness of VisAlign, we evaluate its effectiveness on another state- of-the-art LVLM. We adopt the official Open-Qwen2VL \citep{openqwen2vl} training pipeline , a fully open-source multimodal model that tightly integrates visual embeddings with token-level language representations through shared attention layers. Unlike earlier LVLMs that simply append visual tokens to a frozen LLM, Open-Qwen2VL fuses visual features directly into the token embedding space, enabling richer cross-modal interactions and stronger native visual grounding. In our setup, we follow the prescribed two-stage procedure: we complete Stage-1 visual–language alignment pretraining in full, and then continue the full multimodal pretraining only up to 5000 steps due to compute constraints, using this 5000-step checkpoint as the base model for all subsequent VisAlign experiments.\\\\
We include Open-Qwen2VL as an additional baseline because it represents a more recent and architecturally distinct family of LVLMs, designed specifically for compute-efficient training on academic resources. Its early and tight fusion of visual information with language tokens provides a complementary testbed to Video-LLaVA, allowing us to examine whether VisAlign remains effective when the underlying model already incorporates stronger visual–text coupling. As shown in Table~\ref{tab:llava}, VisAlign consistently enhances performance and reduces hallucinations when integrated into this baseline as well, indicating that rebalancing attention at the embedding level is beneficial even for modern LVLMs with improved native visual grounding.

% \begin{table}[t]
% \centering
% % \small
% \resizebox{0.45\textwidth}{!}{\begin{tabular}{lcccc}
% \toprule
% \textbf{Model} & \textbf{Acc} & \textbf{Precision} & \textbf{Recall} & \textbf{F1-Score} \\
% \midrule
% OpenQwen2VL  (\%)     & 53.13  & 80.12 & 8.33 & 15.09 \\
% + VisAlign    (\%)     & \textbf{55.7} & 56.8  & \textbf{47.6} & \textbf{51.8} \\
% \bottomrule
% \end{tabular}}
% \caption{Effects of VisAlign on the Open-Qwen2VL baseline, on the POPE-AOKVQA benchmark}
% \label{tab:qwen}
% \end{table}

\subsection{Quantitative Assessment of Visual Contribution} \label{appx:quant_vis_contri}

We quantitatively assess “visual contribution” by calculating the proportion of attention allocated to visual tokens (keys) averaged across tokens (queries) and attention heads for 100 randomly selected samples. This analysis is performed only for the first attention layer, as accurate visual-contribution tracking becomes infeasible in deeper layers.

In most LVLMs, visual tokens are inserted in the middle of text tokens, splitting them. Due to the autoregressive nature of the LLM, we skip the first text-token block, since their visual contribution would always be zero.

\begin{table}[t]
\centering
\caption{Visual Contribution (\%) — Baseline vs.\ VisAlign. Results clearly favor VisAlign.}
\label{tab:visual_contribution}
\begin{tabular}{lcc}
\toprule
\textbf{Dataset} & \textbf{Baseline} & \textbf{VisAlign} \\
\midrule
POPE-AOKVQA   & 63.87 & 72.26 \\
POPE-MSCOCO  & 62.90 & 72.68 \\
TextVQA      & 59.69 & 72.27 \\
MME          & 63.00 & 72.89 \\
MM-Vet       & 63.88 & 72.46 \\
\bottomrule
\end{tabular}
\end{table}

\subsection{Additional qualitative results: } 
\begin{figure*}[t]
    % \centering
    \includegraphics[width=\linewidth]{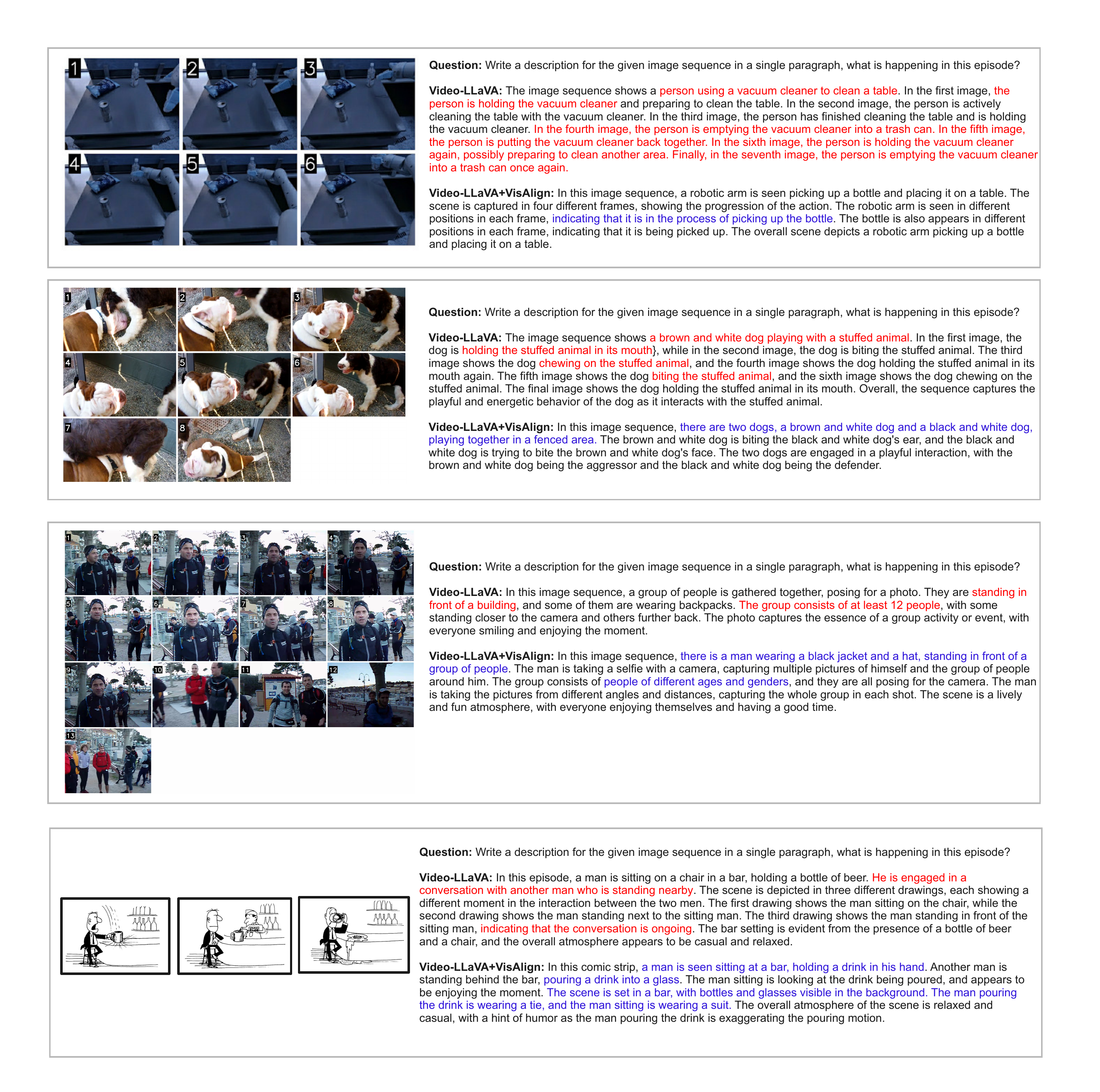}
\caption{Qualitative results on the Mementos benchmark~\citep{wang2024mementos}. Text highlighted in red indicates hallucinated content, while text in blue shows the corresponding corrections. }
    \label{fig:appx_momentos}
\end{figure*}

Figure~\ref{fig:appx_momentos} (see next page) presents additional qualitative results on the Mementos dataset~\citep{wang2024mementos}. As illustrated, VisAlign enables the model to produce more visually grounded predictions and significantly reduces hallucinations compared to the baseline Video-LLaVA model. 
% \label{appx:mementos examples}
% adding a single fig for all examples 

\end{document}